\documentclass[10pt]{article}
\pdfoutput=1
\usepackage{graphicx,graphics}
\usepackage[square,numbers]{natbib} 
\DeclareGraphicsExtensions{.pdf,.png,.jpg}
\usepackage{xspace,xcolor}
\usepackage{fullpage}

\usepackage{amsmath}
\usepackage{amsthm}
\usepackage{amsfonts}

\usepackage{algorithm}
\usepackage{algorithmicx}
\usepackage[noend]{algpseudocode}

\usepackage{amsmath,amsfonts}
\usepackage{amssymb,amsopn}
\usepackage{bm} 

\usepackage{amssymb}
\usepackage{amsmath,amsfonts}
\usepackage{amsthm,amsopn}

\usepackage{bm} 

\newcommand{\vct}[1]{\boldsymbol{#1}} 
\newcommand{\mat}[1]{\boldsymbol{#1}} 
\newcommand{\cst}[1]{\mathsf{#1}}  

\newcommand{\field}[1]{\mathbb{#1}}
\newcommand{\R}{\field{R}} 

\newcommand{\ProbOpr}[1]{\mathbb{#1}}

\newcommand{\expect}[2]{%
\ifthenelse{\equal{#2}{}}{\ProbOpr{E}_{#1}}
{\ifthenelse{\equal{#1}{}}{\ProbOpr{E}\left[#2\right]}{\ProbOpr{E}_{#1}\left[#2\right]}}} 
\newcommand{\var}[2]{%
\ifthenelse{\equal{#2}{}}{\ProbOpr{VAR}_{#1}}
{\ifthenelse{\equal{#1}{}}{\ProbOpr{VAR}\left[#2\right]}{\ProbOpr{VAR}_{#1}\left[#2\right]}}} 

\newcommand{\vc}{\vct{c}}
\newcommand{\vi}{\vct{i}}

\newcommand{\vs}{\vct{s}}

\newcommand{\vo}{{\vct{o}}}

\newcommand{\vr}{\vct{r}}
\newcommand{\vS}{\vct{S}}
\newcommand{\vI}{\vct{I}}

\newcommand{\vv}{\vct{v}}
\newcommand{\vw}{\vct{w}}

\newcommand{\mA}{\mat{A}}
\newcommand{\mB}{\mat{B}}
\newcommand{\mF}{\mat{F}}
\newcommand{\mW}{\mat{W}}

\newcommand{\mS}{\mat{S}}

\newcommand{\mT}{\mat{T}}

\newcommand{\mP}{\mat{P}}

\newcommand{\cM}{\cst{M}}

\newcommand{\cT}{\cst{T}}

\newcommand{\cR}{\cst{R}}

\newcommand{\cW}{\cst{W}}

\newcommand{\mR}{\mat{R}}

\newcommand{\mI}{\mat{I}}

\newcommand{\vf}{\vct{f}}
\newcommand{\vh}{\vct{h}}
\newcommand{\vg}{\vct{g}}

\newcommand{\eat}[1]{}
\usepackage{hyperref}
\usepackage{url}
\newcommand{\ie}{\emph{i.e.}}

\begin{document}

\title{Aligning where to see and what to tell: image caption with region-based attention and scene factorization}

\author{Junqi Jin$^{1\dagger}$, Kun Fu$^{1\dagger}$, Runpeng Cui$^{1}$, \\
Fei Sha$^{2\P}$, Changshui Zhang$^{1\P}$\\[0.5em]
$^1$ Dept. of Automation, Tsinghua University, Beijing, 100084, China\\
 \texttt{\{jjq14@mails, fuk11@mails, crp13@mails, zcs@mail\}.tsinghua.edu.cn}\\[0.5em]
$^2$ Dept. of Computer Science,  U. of Southern California, Los Angeles, CA 90089\\
  \texttt{feisha@usc.edu}\\[0.5em]
$^\dagger$: performed this research while visiting USC and contributed equally\\[0.5em]
$^\P$: to whom questions and comments should be sent
}

\maketitle

\begin{abstract}
Recent progress on automatic generation of image captions has shown that it is possible to describe the most salient information conveyed by images with accurate and meaningful sentences. In this paper, we propose an image caption system that exploits the parallel structures between images and sentences. In our model, the process of generating the next word, given the previously generated ones, is aligned with the visual perception experience where the attention shifting among the visual regions imposes a thread of visual ordering. This alignment characterizes the flow of  ``abstract meaning'', encoding what is semantically shared by both the visual scene and the text description. Our system also makes another novel modeling contribution  by introducing scene-specific contexts that capture higher-level semantic information encoded in an image. The contexts adapt language models for word generation to specific scene types. We benchmark our system and contrast to published results on several popular datasets. We show that using either region-based attention or scene-specific contexts improves systems without those components. Furthermore, combining these two modeling ingredients attains the state-of-the-art performance.
\end{abstract}

\section{Introduction}
\label{sIntro}
The recent progress on automatic generation of image captions has greatly disrupted the well-known adage that \emph{a picture is worth a thousand words}.  Granted, it is still reasonable to assert that an image contains a vast amount of visually discernible information that is difficult to be completely characterized with natural languages. Nonetheless, many image captioning systems have shown that it is possible  to describe the most salient information conveyed by images with  accurate and meaningful sentences~\cite{donahue2014long,fang2014captions,karpathy2014deep,mao2014explain,vinyals2014show,xu2015show}.  

Those systems  have attained very promising results by leveraging several crucial advances in computer vision and machine learning: optimizing on curated datasets of a large number of images and their corresponding human-annotated captions~\cite{chen2015microsoft}, representing images with rich visual features  designed for related tasks such as object recognition and localization~\cite{krizhevsky2012imagenet, simonyan2014very}, and learning highly complex models that are capable of  generating human-readable sentences~\cite{bahdanau2014neural,hochreiter1997long,sutskever2011generating}.

Despite the progress, image caption remains a challenging task. In its most abstract form, the caption algorithm needs to infer the most likely sentence, in the form of a sequence of words $\vS$ given an image $\vI$ --- so far, the most common approach is to define the inference process with a condition probability model $P(\vS|\vI)$. However, specifying the exact form of this model entails careful tradeoff of several design decisions: how to represent the images, how to represent the sentences (\ie, language modeling) and how to fuse the visual information with the textual information, encoded by the images and the sentences respectively.
\begin{figure}[t]
\centering
\small
\includegraphics[width=0.95\columnwidth]{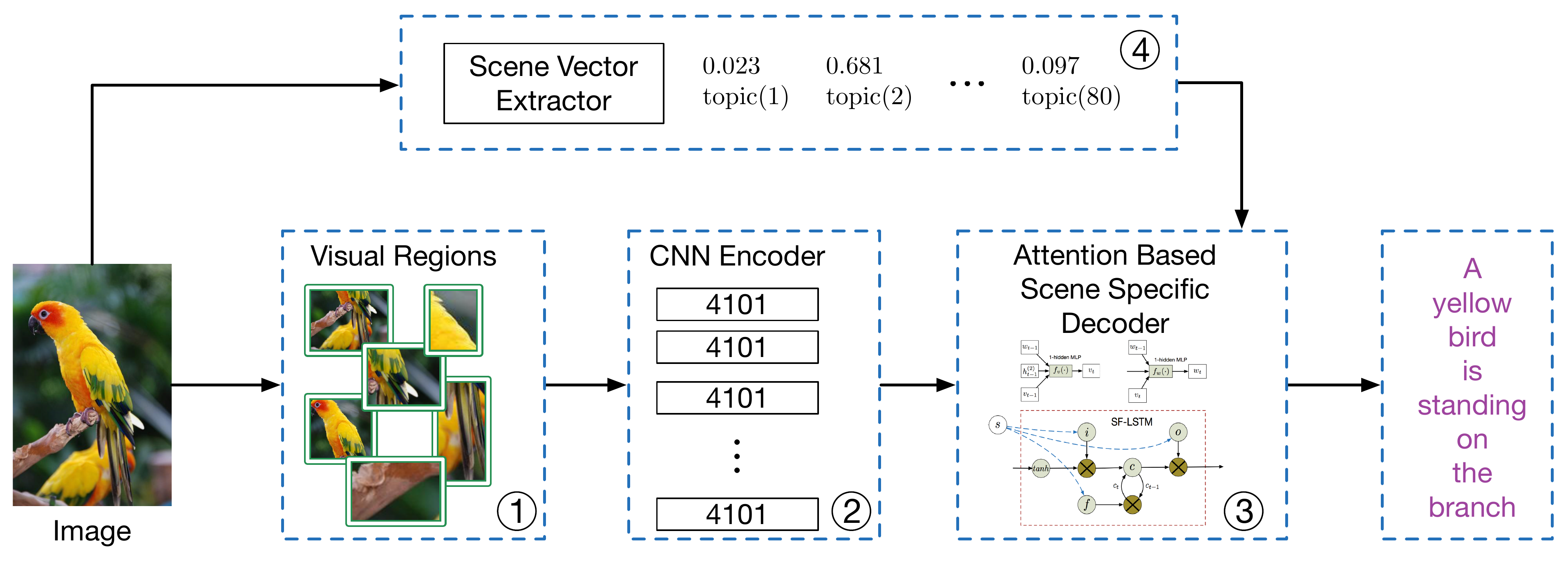}
\caption{\small The architectural diagram of our image caption system. An image is first analyzed and represented with multiple visual regions from which visual features are extracted (section~\ref{sPatch}). The visual feature vectors are then fed into a recurrent neural network architecture which  predicts both the sequence of focusing on different regions  and the sequence of generating words based on the transition of visual attentions (section~\ref{sLSTM}). The neural network model is also governed by a scene vector, a global visual context extracted from the whole image. Intuitively, it selects a scene-specific language model for generating texts (cf section~\ref{sScene}).}
\label{fig:conceptual}
\end{figure}

Arguably, the starting point for an image caption system is to understand the image, for instance, recognizing the objects in it, reasoning the relationship among those objects, and focusing on the more salient parts in the image. The identified objects correspond to the nouns in the caption to be generated, and the relationship corresponds to other linguistic constituents (such as verbs) and determines how to sequentially order the words into a sentence. Finally, the desiderata to keep only the most salient information  tunes out secondary information in the generated caption.

In this paper, we propose an image caption system that follows this modeling idea, and exploits the parallel structures between images and sentences. Fig.~\ref{fig:conceptual} shows the conceptual diagram of our system.  Specifically, we assume that there is a close correspondence between  visual concepts -- detected as object-like regions,  and their textual realization as words in sentences.  Moreover, the process of generating the next word, given the previously generated ones, is aligned with the visual perception experience where the attention shifting among the regions imposes a thread of visual ordering. This alignment characterizes the flow of a latent variable of ``abstract meaning'', encoding what is semantically shared by  both the visual scene and the text description, and is modeled with a recurrent neural network. The hidden states of this network are thus used to predict both where the next visual focus should be  and what the next word in the caption should be.

Our work also introduces another novel modeling contribution with  scene-specific contexts. Such contexts capture higher-level semantic information encoded in an image, for example, the places the image is taken and the possible activities of involving the people in the image. They adapts  language models for generating words to specific scene types. For instance, it is unlikely to caption an image as ``Mary is asleep'' if the scene is about kitchen. Rather a more likely caption would be ``Mary lies on the floor''.  The scene contexts are extracted visual feature vectors from the whole image and affect the word generation by biasing the parameters in the recurrent neural network.  

Our systems differ from others in the following ---  detailed comparisons are deferred until after describing ours. We identify localized regions at multiple scales, which contain visually salient objects, to represent images. Those regions ground the concepts in sentences.  Systems such as ~\cite{vinyals2014show,donahue2014long,mao2014explain} extracts visual features from  images as a whole, thus are unable to provide fine-grained modeling of the interdependencies of different visual elements. \cite{karpathy2014deep,fang2014captions} models images as a collection of patches. However, their two-stage systems focus on learning the mapping between the visual patches and the words. The detected words for test images are then used by language models to generate sentences. Our system not only resolves the correspondence between the regions and the words but also models the parallel transitioning dynamics between the visual focus  and the sentences.

We benchmark our system and contrast to published results by others on several popular datasets. We show that using either region-based attention or scene-specific contexts improves systems without those components. Furthermore, combining these two modeling ingredients attains the state-of-the-art performance, outperforming other competing methods by a noticeable margin.

The rest of the paper is organized as follows. We describe  related work in sec.~\ref{sRelated}, followed by a detailed description of our system in sec.~\ref{sApproach}. We report empirical results in sec.~\ref{sExp} and conclude in sec.~\ref{sConclusion}.

\section{Related Work}
\label{sRelated}
Image caption generation has long been a challenging problem in computer vision. A traditional approach is to use pre-defined templates to generate sentences by filling detected visual elements such as objects \cite{kulkarni2013babytalk,li2011composing,yang2011corpus,mitchell2012midge,elliott2013image}. Retrieval based models \cite{kuznetsova2012collective,kuznetsova2014treetalk} first find a similar image in training set, and compose a new sentence based on the retrieved images' sentences. These methods' generated sentences are very fixed and limited, and cannot describe specific contents in a test image.

Recent work aims to automatic generating words with language models learnt from data.  \cite{kiros2014unifying} proposed a multi-modal log-bilinear model to generate sentences with a fixed context window.  \cite{mao2014explain} proposed a multimodal Recurrent Neural Network (m-RNN) architecture to fuse text information and visual features extracted on the whole image.   \cite{vinyals2014show} used a deep CNN to extract the whole image feature, and the feature is only input once to an RNN as the initial start word.  \cite{fang2014captions} first used multiple instance learning to train a word detector. Given a novel test image, they used the detector to detect words from patches extracted from the image. Then they used a maximum entropy language model to generate the sentence with the detected words, and at last, they re-rank the generated sentences with minimum error rate training. \cite{karpathy2014deep} aligns words in sentence with patches in an image, in which images are represented as CNN features computed on patches and words are embedded using a bidirectional RNN. But during testing, the whole image is used to extract a visual context and supply to language models. 

\cite{xu2015show} is closest to us in spirit. It represents images with features computed on tiles, fix-sized regions. They then learn a neural network to predict the changing locations of the attention and generate word based on the located tile. \cite{donahue2014long} proposed to use multiple layers of LSTMs and investigate the best way to feed visual and text information to different LSTMs. Our system combines many features in the existing systems: we use localized regions at multiple scales to represent images, multiple layers of LSTMs to encode the evolvement of ``abstract meaning'' shared by the visual and textual information, and learn end-to-end to adjust the system components to generate the desired sentences.
\section{Approach}
\label{sApproach}

Our system for image caption is composed of the following components: visual features representation of the image with localized regions at multiple scales (section~\ref{sPatch}), an LSTM-based neural network that models the attention dynamics of focusing on those regions as well as generating sequentially the words (section~\ref{sLSTM}), a visual scene model that adjusts the LSTM to specific scenes (section~\ref{sScene}). We describe in detail each of those components, followed by describing the numerical optimization procedures and other  details such as comparison to related work. 

\subsection{Image representation with localized patches at multiple scales}
\label{sPatch}

\begin{figure}[t]
\centering
\includegraphics[width=0.95\columnwidth]{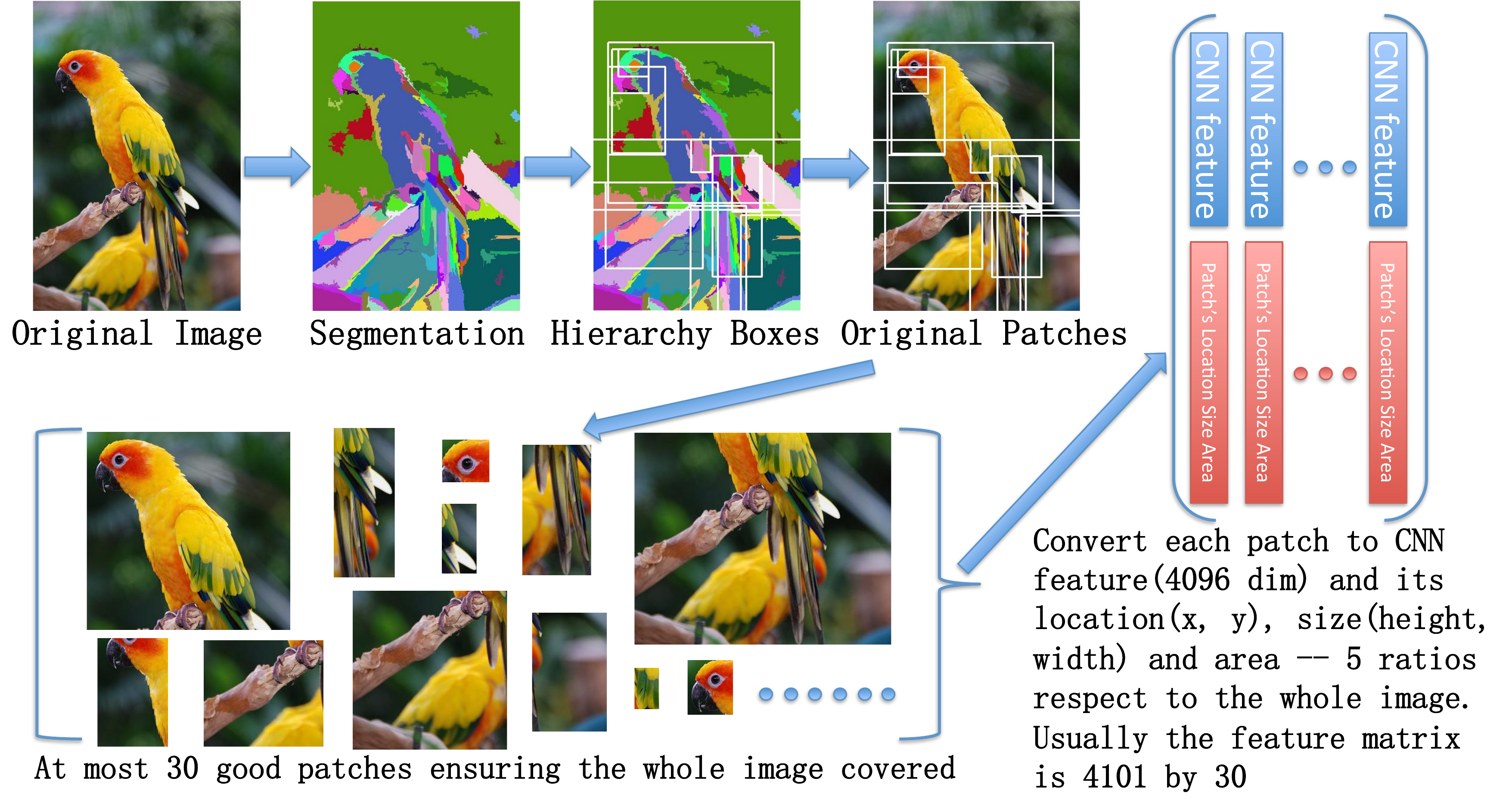}
\caption{\small Image representation with localized regions at multiple scales. The image is  hierarchically segmented  and top regions containing  salient visual information are selected.  Those regions are  analyzed with a Convolutional Neural Net (CNN) trained for object recognition. Resulting visual features, along with the spatial  and geometric information are concatenated as feature vectors. See texts for details. }
\label{fig:selectiveSearch}
\end{figure}

In stark contrast to many existing work that represent the image with a global feature vector~\cite{mao2014explain,vinyals2014show,donahue2014long}, our system represents the image as a collection of feature vectors computed on localized regions at multiple scales. This representation provides an explicit grounding of concepts (words in the sentence) to the visual elements in the image. It also enables fine-grained  modeling how those concepts should be pieced together --- with an attention model --- to be described in the next section.  Fig.~\ref{fig:selectiveSearch} illustrates the main steps.

\paragraph{Generate candidate regions/patches} We use the technique of selective search~\cite{uijlings2013selective} to construct a hierarchical segmentation of the image.  The technique first uses color and texture features to over-segment the image, and merge neighboring regions to form a hierarchy of segmentations until the whole image merge into a single region (for clarity, Fig. \ref{fig:selectiveSearch} only shows the lowest level segmentation). For each identified region, a tight bounding box is used to delineate its boundaries.

\paragraph{Select good visual elements} Among the vast collection of regions from different levels (ie, scales), we select  ``good'' ones as the first means of focusing on the most salient and relevant visual elements to be captioned. We define ``goodness'' in the following desiderata:  (1) semantically meaningful: those regions  should contain high-level concepts that can be described by natural language phrases. (2) primitive and non-compositional: each region should be small enough to contain a single concept that can be captured with words, short phrases etc. (3) contextually rich: each region should be large enough that it contains neighboring contextual information such that visual features extracted from the region can be indicative of inter-dependency of other visual elements in the image.  Note that the goals of (2) and (3) are naturally in conflict and need to be carefully balanced.

To this end, we train a  classifier to learn whether the region is good or bad --- details for constructing this classifier are in the Appendix.  For each image, we select the top $\cR=30$ regions, according to the outputs of the classifier, under the constraint that their union should cover the whole image and the sizes of the bounding boxes for the regions are diverse. Diverse sizes are preferred as not all objects in a image have the same size. Moreover, abstract concepts such as verbs tend to be strongly associated with heterogeneous scales --- for example, ``flying the sky'' might needs a very large patch to be recognized while ``red'' color can be decided on much smaller scales. We achieve the selection of diverse sizes by randomly permuting the ranking order of the scores.

\paragraph{Extract visual features from regions} We resize each patch into 224$\times$224 to feed into 16-layer VGG-net \cite{Simonyan14c} to obtain 4096-dimensional CNN features. We also add the box's center's x location, y location, width, height and area ratios with respect to the whole image's geometry.

\paragraph{Comparison to other systems using regions}  While detecting objects in the image, \cite{fang2014captions,karpathy2014deep} focus on deriving the latent alignment between the detected regions and the words in the training sentences. Their purpose is to use the alignments to train a recurrent neural network generator of word sequences where the training data have become the aligned regions and corresponding words. However, when captioning a new image, the trained generator takes the feature vector computed over the whole test image as in other similar systems~\cite{mao2014explain,vinyals2014show}. In contrast, we do not need to explicitly learn the alignments and we use image regions on test images.

\cite{xu2015show} is closest to our system in spirit. The authors there define same-sized, totally $14\times 14$ regions using grids and represents images at finer scale with the collection of feature vectors extracted from those regions. Note that due to the pre-determined size, their regions could either correspond to a partial view of a visual element (ie, a concept) or a conglomeration of several concepts (if the region is too big).  Both their system and ours leverage the intuition that different parts of sentences ought to correspond to different regions on the image. To this end, both systems need to model how captioning moves between regions, using an attention model to characterize the dynamics. We will describe this model in the next.

\subsection{Attention-based Multi-Modal LSTM  Decoder}
\label{sLSTM}
\begin{figure}[t]
\begin{center}
   \includegraphics[width=0.9\columnwidth]{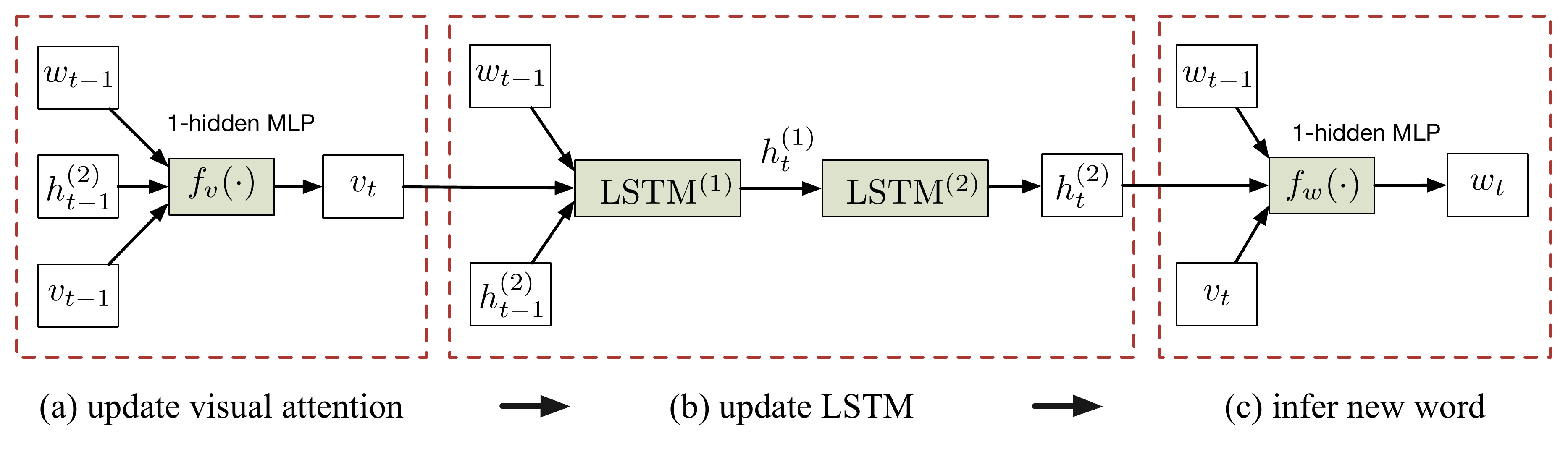}
\end{center}
   \caption{\small The parallel processes of generating a new word and shifting visual attentions among visual regions on the image. (a) Based on the history of previously generated word, hidden states that encode the flow of the abstract meaning, and the previous visual attention, a neural network predicts which visual region should be focused on now. (b) Based on the new focus, the hidden states are updated. (c) the new hidden states, together with the previous word, and the current visual features, predicts which word will be generated next. }
\label{fig:framework}
\end{figure}

When we visually perceive a set of visual elements scattering on the 2-D image, how does our cognitive process generate a sentence which is a sequentially structured linear chain of words to describe it?  We hypothesize there is a latent process  $\{\vh_t\}$ of ``abstract meaning'', governing the transitions from one concept to another. When this process is used to drive the generation of words, it yields a textual form of the abstract meanings encoded in the image. When this process is used to analyze the set of visual elements, it gives rise to a trajectory of visual attention, directing what and where our visual perception system should attend to. Our modeling is thus inspired by the recent work~\cite{xu2015show};  we postpone the comparison to later, after explaining the architecture of our model.

\paragraph{Overview} In our model, we have 3 sets of variables:  the hidden states $\{\vh_t\}$ for characterizing the transition of abstract meanings, the output variables $\{\vw_t\}$ for the words being generated, and  the input variables $\{\vv_t\}$ describing the visual context for the image, for example, for the visual element(s) being focused. For simplicity, the subscript $t$ indices the ``time'', expanding from 0 (\textsc{start}) to $\cT+1$ (\textsc{end}) where $\cT$ is the length of the sentence. 

The output variable $\vw_t \in \R^{\cW}$ is an one-hot column vector in the form of 1-of-$\cW$ encoding scheme.  $\cW$ is the number of words in the vocabulary and $\vw_t$ has only one element being 1 and  all other elements being 0. We learn an embedding matrix $\mP_{\vw}$ to convert $\vw_t$ into a point in $\cM$-dimensional Euclidean space ($\cM= 256, 512$ and $512$ for Flickr8K, Flickr30K and MSCOCO respectively).

Our model is a sequence model, predicting the value of the new state and output variables at time $t+1$, based on their values in the past as well as the values of input variables up to time $t+1$. Of particular importance, is that the dynamics of $\vh_t$ is modeled with a LSTM unit. We explain how those predictions are made in the following. 
Fig.~\ref{fig:framework} gives an overview of those predictive models.

\paragraph{Predict visual focus} At any time $t$, our system is presented with the image which is already analyzed and represented with $\cR$ localized patches at multiple scales (cf. section~\ref{sPatch}). We denote the collection of the feature vectors computed from those regions as $\mR = \{\vr_1, \vr_2, \ldots, \vr_{\cR}\}$.

At time $t$, we predict which  visual element is being focused and obtain  the right feature vector as visual context. We use an one-hidden-layer neural network with $\cR$ softmax output variables:
\begin{equation}
p_{it} \propto \exp\,\{ f_v(\vr_i, \mP_{\vw}\vw_{t-1}, \vh_{t-1}, \vv_{t-1})\}, \forall\ i = 1, 2, \cdots, \cR
\end{equation}
where $p_{it}$ denotes the probability of focusing on $i$-th region at time $t$. $f_v(\cdot)$ parameterizes the neural network mapping function (before the softmax). Note that the inputs to the neural network include the extracted features from the $i$-th visual element, and the histories at the previous time step, including the generated word, the abstract meaning  and the visual  context $\vv_{t-1}$. The parameters of the neural network are learnt from the data.

To select the visual element to focus on, we can sample a visual element $r$ based on the probabilities $\{p_{it}\}$ and assign the corresponding $\vr_r$ as the current visual feature context $\vv_t$. A simpler approach is to just update the visual feature context using weighted sum 
\begin{equation}
\vv_t = \sum_{i} p_{it} \vr_i
\end{equation}
We adopt the simpler weighted sum, though both mechanisms were studied by~\cite{xu2015show}.  Note that if the outputs of the softmax layer are highly peaked around a particular visual element, then the weighted sum approximates well the feature element of the visual element on which is to be focused. 
We use 256 hidden units for Flickr8K, and 512 for Flickr30K and MSCOCO.

\paragraph{Update meaning trajectory} Given the newly predicted visual context $\vv_t$, we update the abstract meaning $\vh_t$. This is a crucial component in our model and we have used two LSTM units stacked up to model the complex dynamics of $\vh_t$, illustrated in Fig.~\ref{fig:framework}.

The input to the bottom LSTM is the tuple $\mP_{\vw}\vw_{t-1}, \vh_{t-1}$ and $\vv_t$, composed of both the histories and the new visual context. The output of the bottom LSTM is then fed into the top LSTM as the input. The output of the top LSTM is $\vh_t$. The equations describing the LSTMs are in the Appendix.  For both layers, memory cells have the same size as hidden layers.

\paragraph{Predict next word} Given the updated abstract meaning $\vh_t$, we predict the next word $\vw_t$ with an one-hidden-layer neural network with $\cW$ softmax output units. Specifically,
\begin{equation}
p_{wt} \propto \exp\,\{ f_w(\mP_{\vw}\vw_{t-1}, \vh_{t}, \vv_{t})\}, \forall\ w = 1, 2, \cdots, \cW
\end{equation}
where $f_w$ parameterizes the neural network mapping function up to the softmax layer.

\paragraph{Other details and comparison to other decoder methods}  $\vv_0$ is initialized as the averaged region features. The LSTMs' $\vc^{(1)}_0$, $\vh^{(1)}_0$, $\vc^{(2)}_0$, $\vh^{(2)}_0$ are initialized using 4 independently trained  MLPs which take $\vv_0$ as input and have 1 hidden layer with the same size as $\vv_0$.

Our approach is directly inspired by the attention-based decoder in ~\cite{xu2015show}, though we have modified slightly to use two stacked LSTM layers as well as incorporating the previous visual context in order to predict the next visual element to be focused.

\subsection{Scene Factored LSTM}
\label{sScene}

\begin{figure}[t]
\centering
   \includegraphics[width=0.85\columnwidth]{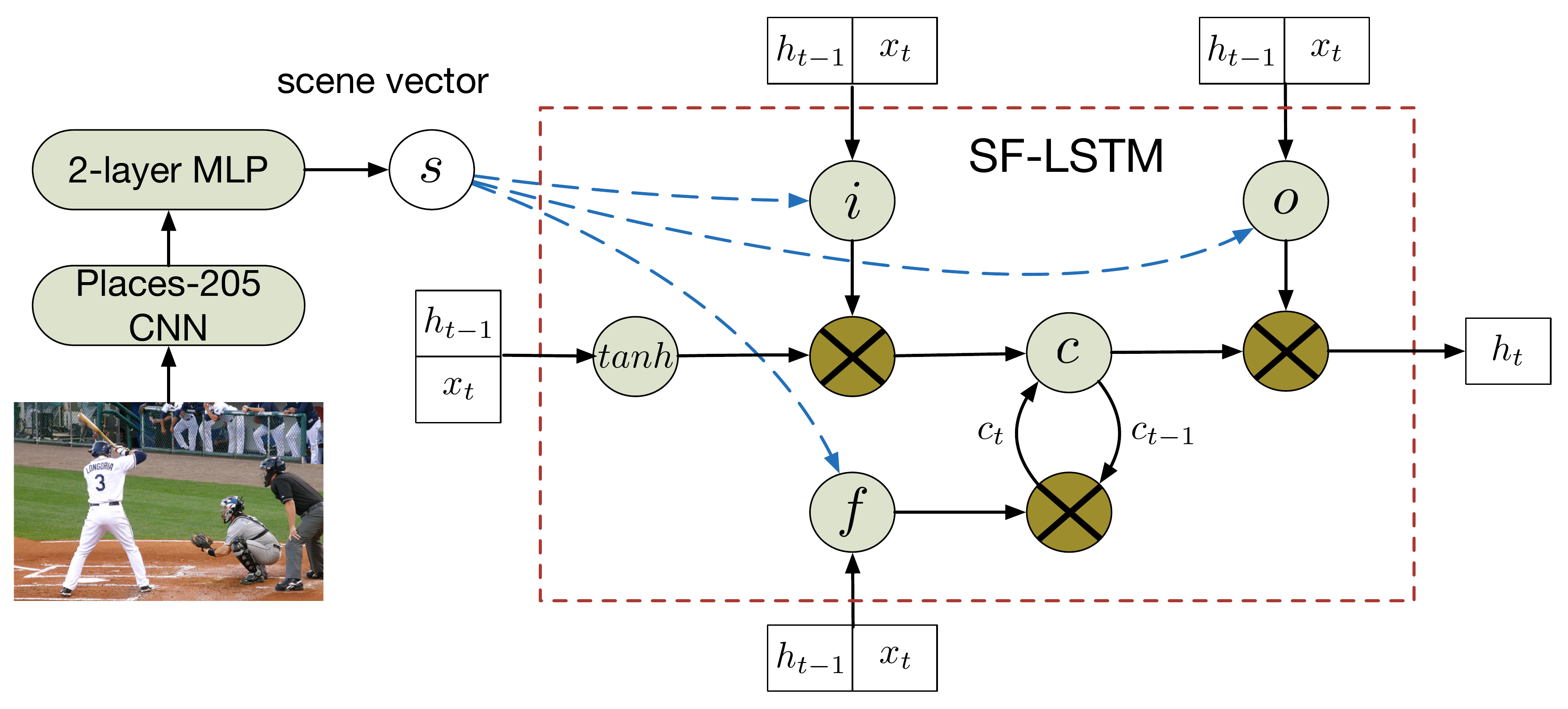}
   \caption{\small The scene vector $s$ is extracted by Places-205 CNN. The right part is the basic unit of LSTM. $s$ is used to factor the weight matrix in the 3 gates.}
\label{fig:LSTM}
\end{figure}

Imagine a photo where a person walks along with a dog. If the   photo is taken at a park, a likely caption could be \emph{A person is walking the dog for exercising}. On the other hand, if the photo is taken at a pet store, a more likely caption could be \emph{A person takes the dog to be groomed}. Intuitively, the scene category is an invaluable context that can affect the selection of words significantly --- the word \emph{groom} should not appear in the park scene, while the word \emph{exercise} is unlikely a typical activity in a pet store and thus should not be selected with high confidence.

How can we exploit such global contexts for better image captioning? To this end, we describe another contribution by our work: scene-factored LSTM. We describe first how to extract scene-related global contexts, followed by how to inject scene contexts into LSTMs.

\paragraph{Scene-specific contexts} Our goal is to obtain a scene vector for each image. For the purpose of using this vector for better captioning,  this scene vector should be informative of textual descriptions  and also needs to be inferable from visual appearance. We achieve these goals with two steps: unsupervised clustering of captions into ``scene'' categories and supervised learning of a classifier to predict the scene categories from the visual appearance.

For the first step, we use Latent Dirichlet Allocation(LDA)~\cite{blei2003latent} to model the corpus of all the captions in the training dataset of MSCOCO~\cite{lin2014microsoft}. For the second step, we train a multilayer perceptron to predict the topic vector, computed by LDA,  from each image's visual feature vector. Note that this predictive model allows to extract topic vectors for images without captions. We call the topic vectors as scene vectors. Details are in the Suppl. Material.

\paragraph{Adapt LSTMs to be scene-specific} The LSTMs (as described in section~\ref{sLSTM}) encodes the language model how the words should be sequentially selected from the vocabulary. To inject scene vectors and thus adapt the sentence generation process to be scene-specific, we factorize the parameters in the LSTMs. Specifically, given an image and its associated scene vector $\vs$, we use ``personalized'' LSTMs for that image to generate caption. Concretely, for all gates, the affine transformations will be reparameterized as Fig.~\ref{fig:LSTM}. To avoid notation cluttering, assume we have a linear transformation matrix $\mW$ to be applied to $\vh_{t-1}^{(1)}$. Then $\mW$ is given by
\begin{equation}
\mW = \mA\, \mathsf{diag}(\mF\vs)\, \mB
\end{equation}
where $\mA$ and $\mB$ are two (suitably-sized) matrices that are \emph{shared} by all scenes (and all images). $\mF$ is another matrix that linearly transforms the scene vector $\vs$. This technique has been previously used in \cite{taylor2009factored,sutskever2011generating} where they factorize movement style and language models. The size of the $\mF$ is $512\times 80$ for Flickr8K, $1024 \times 80$ for Flickr30K and MSCOCO. 

Note that our design of the scene-specific LSTMs represent a tradeoff between specialization and the number of parameters to learn. Another option is to categorize each image into a ``hard'' scene assignment (in lieu of our current soft scene membership assignment due to $\vs$ being a probability vector, representing the mixture of different scene categories), and then for each scene, learn a set of LSTMs parameters that do not share among them (instead of the sharing of $\mA$ and $\mB$ in our factorization scheme). The drawback of this option is that the number of parameters to be learned will increase linearly with respect to the number of scenes. Additionally, the language models learned by different LSTMs will not share any commonness --- this might not be desirable as there are certainly many scene-independent language components.

\section{Experiments}
\label{sExp}

We evaluate our image caption system on several datasets and compare to several state-of-the-art systems on both the task of captioning and the image/text retrieval tasks.  We also evaluate our system qualitatively by validating our modeling assumptions of region-based attention and scene-factorization for language models. We describe the setup of our empirical studies first, followed by discussing both quantitative and qualitative results.

\subsection{Setup}

\paragraph{Datasets and evaluation metrics} We have experimented on 3 datasets: MSCOCO~\cite{lin2014microsoft}, Flickr8K~\cite{rashtchian2010collecting} and Flickr30K~\cite{young2014image}.  We have followed the standard protocols to split data into training, validation and evaluation. Details are in the Appendix.   We evaluate captions on test images in the following metrics: BLEU-1, BLEU-2, BLEU-3, BLEU-4, METEOR, ROUGE-L and CIDEr-D, given by MSCOCO server~\cite{chen2015microsoft}. In essence, they measure the agreements between the ground-truth captions and the outputs of automatic systems.  We use the public Python evaluation API  provided by the MSCOCO server. 

\paragraph{Alternative methods} We compare to several recently proposed image captioning systems: DeepVS \cite{karpathy2014deep}, mRNN \cite{mao2014explain}, Google NIC \cite{vinyals2014show}, LRCN \cite{donahue2014long}. All those systems have published publicly available evaluation results on the 3 databases we have experimented. While our work is inspired by~\cite{xu2015show}, their system's evaluation procedures are different from ours and other systems\footnote{According to those authors, ``we report BLEU from 1 to 4 without a brevity penalty''.}. 

\paragraph{Implementation details} To generate a caption, one needs to sample the word to be output at time $t$. A sound strategy is to form a beam search where a pre-determined number (called ``beam size'') of best-by-now  sentences (up to time $t$) are computed and kept to be expanded with new words in future. In our experiments, the beam size is set as 10. We also experimented the ``greedy search'' where the beam size is set to 1.  Other optimization details are provided in the Appendix.

\subsection{Quantitative evaluation results}

\begin{table}[t]
  \centering
  \small
  \caption{Evaluation of various systems on the task of image captioning, on MSCOCO dataset}
\label{tab:caption}
    \begin{tabular}{|c|ccccccc|}
    \hline
    & BLEU-1 & BLEU-2 & BLEU-3 & BLEU-4 & METEOR & ROUGH-L & CIDEr-D  \\
    \hline
    \hline
    DeepVS\cite{karpathy2014deep} & 62.5 & 45.0 & 32.1 & 23.0 & 19.5 & -- & 66.0 \\
    LRCN\cite{donahue2014long} & 62.8 & 44.2 & 30.4 & 21.0 & -- & -- & -- \\
    Google NIC \cite{vinyals2014show}& 66.6 & 46.1 & 32.9 & 24.6 & -- & -- & -- \\
    mRNN\cite{mao2014explain} & 67 & 49 & 35 & 25 & -- & -- & -- \\
    \hline
    \textsc{our-base-greedy} & 64.0 & 46.6 & 32.6 & 22.6 & 20.0 & 47.4 & 70.7 \\
    \textsc{our-sf-greedy} & 67.8 & 49.4 & 34.8 & 24.2 & 21.8 & 49.1 & 74.3 \\
    \textsc{our-ra-greedy} & 67.7 & 49.5 & 34.7 & 23.5 & 22.2 & 49.1 & 75.1 \\
    \textsc{our-(ra+Sf)-greedy} & 69.1 & 50.4 & 35.7 & 24.6 & 22.1 & 50.1 & 78.3 \\
    \textsc{our-(ra+sf)-beam} & \textbf{69.7} & \textbf{51.9} & \textbf{38.1} & \textbf{28.2} & \textbf{23.5} & \textbf{50.9} & \textbf{83.8} \\
    \hline
     \end{tabular}
    
  \end{table}

\paragraph{Main results} Table~\ref{tab:caption} compares our method to several other systems on the task of image caption. We consider several systematic variants of our method: (1) \textsc{our-base}  is similar to Google NIC --- we represent the images with CNN features computed on the whole image, stripping away the aspects of region-based attention and scene-factorization of our approach. (2) \textsc{our-sf} adds scene-factorized LSTMs to \textsc{our-base}. (3) \textsc{our-ra} adds region-based attention to \textsc{our-base}. (4) \textsc{our-ra+sf} adds both the scene-factorized LSTMs and the region-based attention to \textsc{our-base}. The terms `-greedy'  and `-beam' denote the search strategies being used.

On the MSCOCO dataset, using the same search strategy of `-greedy', adding either region-based attention (\textsc{our-ra}) or scene-factorized LSTMs (\textsc{our-sf}) improves the base system (\textsc{our-base}) noticeably on all metrics. Moreover, the benefits of region-base attention and scene-factorization are additive, evidenced by the further improvement by \textsc{our-ra+sf-greedy}. By using the beam search, \textsc{our-ra+sf-beam} attains the best performance metrics, outperforming all other competing systems's previously published results.

\paragraph{Other results in the Appendix}  We show that our approach \textsc{our-ra+sf-beam} outperforms other competing methods too (except that our method is close to Google NIC on BLEU-1) on the Flickr8K and Flickr30K datasets. We also demonstrate the state-of-the-art performance by our approaches on the tasks of image/caption retrieval.

\subsection{Qualitative evaluations}

We also evaluate our system qualitatively in two aspects: how the transition dynamics of the abstract meaning $\vh_t$ correspond to the changes of concepts in the caption, and how scene visual contexts influence the generation of the caption.

\paragraph{Trajectory of visual attentions} To illustrate the transition of the attentions, we compute the weighted sum of pixel values in the image. The weights are determined by the amount of attentions predicted for the regions where the pixels belong to. Fig.~\ref{fig:attention} gives an example. We note that first, the weighted sum of pixel values clearly shows the distinction of ``foreground'' (where the attention would be focusing) and the ``background'' where the weighted sum is smallest (ie, darkest). We also observe that the more focused regions (bordered by red rectangles) correspond well to the concepts in the caption. For instance, for ``standing'', the highlighted region contains 4 legs of the cow standing up. For ``grass'', the highlighted region contains mostly the grass (in the background), excluding the cow in the foreground. For other examples, please refer to the Appendix.

\begin{figure}[t]
\centering
   \includegraphics[width=0.95\columnwidth]{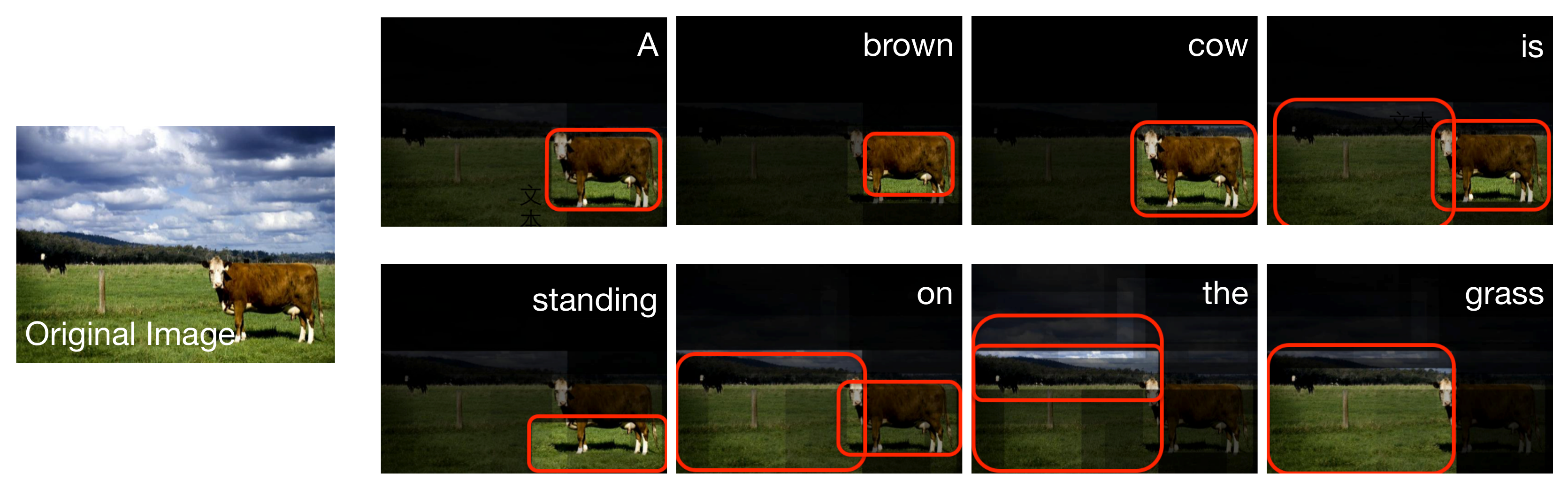}
   \caption{\small The alignment between how attention is shifted among visual regions and the generation of words for the caption. Many  concepts in the sentence correspond well to the visual elements on the image.}
\label{fig:attention}

\end{figure}

\begin{figure}[t]
\centering
   \includegraphics[width=0.95\columnwidth]{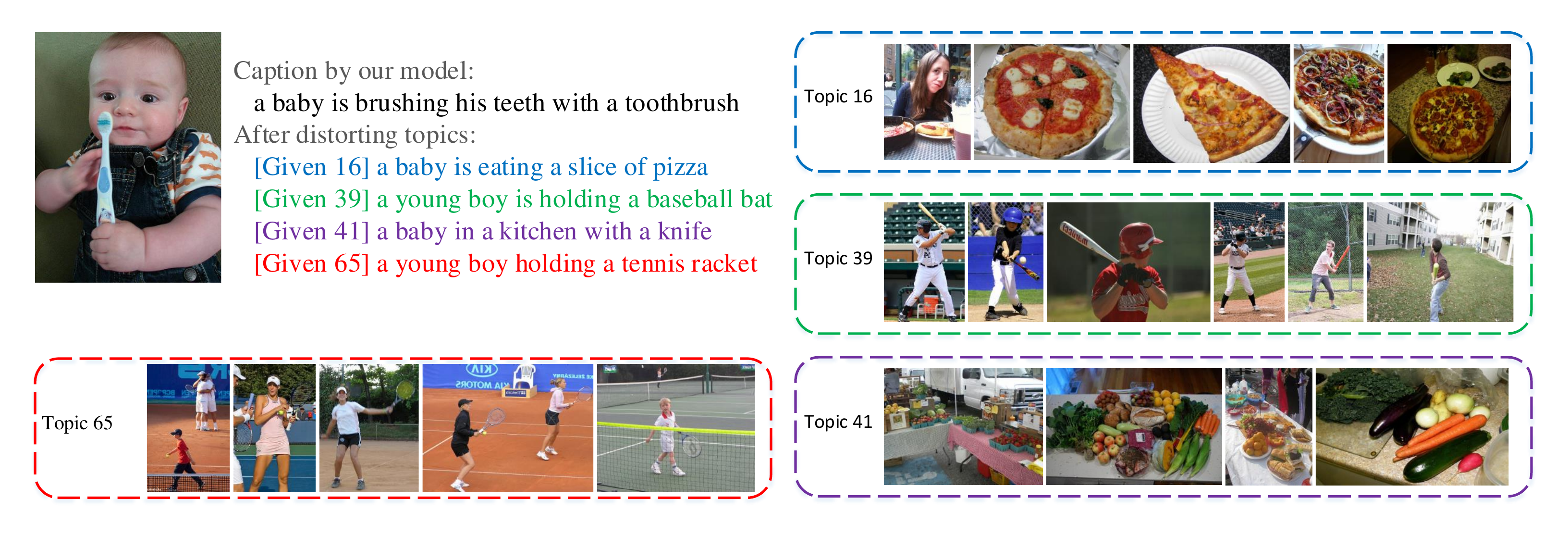}
   \caption{\small  An image from MSCOCO($cocoid=311394$) in which a baby holds a toothbrush. The caption given by our model properly describes the image content by using the correct scene vector to bias the language generation.  To see the influence of scene vectors, we replace the origin scene vector with four 1-hot topic vectors. The topic indexes are 16, 39, 41, and 65. The baby will hold different objects given different scene vectors. An example caption given by model using universal language model can be found in http://cs.stanford.edu/people/karpathy/deepimagesent/.}
\label{fig:sceneAnalysis}
\end{figure}

\paragraph{Effect of scene factors on caption generation} We hypothesize that global visual contexts such as scene categories can significantly affect how captions are generated. We verify this by distorting the predicted scene vector for a test image to another one. We then  use the distorted scene vector to generate a new caption and contrast to the caption obtained from the undistorted scene vector.

Fig.~\ref{fig:sceneAnalysis} exemplifies the effect of using different scene vectors. The  image portrays a baby holding a toothbrush and our model identifies its scene as LDA topic $\#5$.  Distorting the prediction to other topics leads to drastically different captions.  Clearly, topic $\#16$ corresponds to scenes about food, and the caption is now changed to regarding the baby holding a slice of pizza. Similarly, topic $\#65$ is related to the sports scene and the the caption now being changed to the baby holding a baseball bat. 
Those examples demonstrate that scene categories have a significant impact on the caption generation by biasing the language model  to use words that are more common to the targeted scene vector.

\section{Conclusions}
\label{sConclusion}
We propose an image caption system that exploits the parallel structures between images and sentences. One contribution of our system is aligning the process of generating captions and the attention shifting among the visual regions. Another is introducing the scene-factorization LSTM that adapts language models for word generation to specific scene types. Our system is benchmarked and contrasted to published results on several popular dataset including MSCOCO, Flickr8K and Flickr30K. Either region-based attention or scene-specific contexts improves performance. Combining these two modeling ingredients provides a further improvement, attaining the state-of-the-art performance.

\section*{Acknowledgement}

F. S. is partially supported by the Intelligence Advanced
Research Projects Activity (IARPA) via Department of Defense
U.S. Army Research Laboratory (DoD / ARL) contract
number W911NF-12-C-0012, and  additionally  by NSF awards IIS-1065243 and IIS-1451412, a Google Research Award, an Alfred. P. Sloan Research Fellowship and an ARO YIP Award (W911NF-12-1-0241).  The U.S. Government is authorized
to reproduce and distribute reprints for Governmental
purposes notwithstanding any copyright annotation thereon.
The views and conclusions contained herein are
those of the authors and should not be interpreted as necessarily
representing the official policies or endorsements, either
expressed or implied, of IARPA, DoD/ARL, or the U.S.
Government.
C. Z. is partially supported by 973 Program(2013CB329503),  NSFC (Grant No. 91120301, 91420203 and No.61473167)

\bibliographystyle{plain}
\bibliography{ref}

\appendix
\section{Determine objectness of visual elements}
We use the images in the MSCOCO dataset that provides object segmentations. We use the segmented objects as positive examples and un-annotated patches or patches overlapping 20-30\% with positive examples as negative examples. Those patches are resized  to $200\times200$, and represented with HoG features. We then train a logistic regression model to classify the patches and apply it to new images (including those from other datasets). The classifier's outputs is considered as a measure of the objectness of the patches.

\section{LSTM equations}
We found the notation used in \cite{xu2015show} especially elucidating and adapted in the following to describe in detail how our LSTM layers work. We use superscripts $^{(1)}$ and $^{(2)}$ to denote the variables in the bottom and the top LSTM units respectively. We use $\vi_t$, $\vf_t$, $\vo_t$ and $\vc_t$ to represent the outputs of the input, forget, output gates and memory cell. For hidden states, we use $\vh_t^{(1)}$ and $\vh_t^{(2)}$ and set $\vh_t^{(2)} = \vh_t$ as our abstract meaning variable. For the two units, the following equations define how those variables are related:

\begin{equation}
\left(\begin{array}{c}
\vi_t^{(1)}\\
\vf_t^{(1)}\\
\vo_t^{(1)}\\
\vg_t^{(1)}
\end{array}\right) = \left(\begin{array}{c}
\sigma\\
\sigma\\
\sigma\\
\tanh\end{array}\right)\mT^{(1)} \left(\begin{array}{c}
\mP_{\vw}\vw_{t-1}\\
\vh_{t-1}^{(1)}\\
\vh_{t-1}^{(2)}\\
\vv_t
\end{array}\right)
\end{equation}

\begin{equation}
\left(\begin{array}{c}
\vi_t^{(2)}\\
\vf_t^{(2)}\\
\vo_t^{(2)}\\
\vg_t^{(2)}
\end{array}\right) = \left(\begin{array}{c}
\sigma\\
\sigma\\
\sigma\\
\tanh\end{array}\right)\mT^{(2)} \left(\begin{array}{c}
\vh_{t}^{(1)}\\
\vh_{t-1}^{(2)}
\end{array}\right)
\end{equation}

where $\mT^{(1)}$ and $\mT^{(2)}$ are properly defined linear transformation matrices for the two LSTM units.  The sigmoid and the hyperbolic tangent functions $\sigma(\cdot)$ and $\tanh(\cdot)$ are to be applied elementwisely. The memory cell and the hidden state are given by

\begin{equation}
\begin{aligned}
\vc_t^{(1)} &= \vf_t^{(1)} \odot \vc_{t-1}^{(1)} + \vi_t^{(1)}\odot \vg_t^{(1)} \\
\vh_t^{(1)} &= \vo_t^{(1)} \odot \tanh (\vc_t^{(1)})
\end{aligned}
\end{equation}

\begin{equation}
\begin{aligned}
\vc_t^{(2)} &= \vf_t^{(2)} \odot \vc_{t-1}^{(2)} + \vi_t^{(2)}\odot \vg_t^{(2)} \\
\vh_t^{(2)} &= \vo_t^{(2)} \odot \tanh (\vc_t^{(2)})
\end{aligned}
\end{equation}
where $\odot$ stands for element-wise multiplication.  Note that at time $t$, the top LSTM uses the updated hidden states $\vh_t^{(1)}$ from the bottom layer.

%
\section{Scene vector}
Concretely, for the first step, we use Latent Dirichlet Allocation(LDA)~\cite{blei2003latent} to model the corpus of all the captions in the training dataset of MSCOCO. For each image, we obtain a 80-dimensional topic vector that ``softly'' assigns its caption into the memberships of 80 categories.  We call the topic vectors the ``scene vectors''. Note that the scene vectors are purely inferred from captions. For the second step, we train a multilayer perceptron to predict the scene vector when presented with an image. The training samples for this classifier are the images from the same training dataset of MSCOCO with the target outputs being the LDA-inferred scene vectors. We use an MLP with two hidden layers with the sizes of 1024 and 512. We use softmax in the last layer and sigmoid function for others.

We represent the training images with global feature vectors computed on the whole image. While it is possible to use any CNN trained on object recognition tasks, we use the CNN from the Places-205 CNN~\cite{zhou2014places}. Places-205 CNN is based on AlexNet~\cite{krizhevsky2012imagenet}, but optimized under a 2.4 million datatbase to predict the locations of the images. We use the computed features at the outputs of the last fully-connected layer.

Note that, representing images with global feature vectors and using the scene classifier provide an effective way to categorize test images where captions are not available (thus scene vectors cannot be inferred from LDA). Specifically, when generating captions for new images, scene vectors predicted from MLP are used.

%
\section{Empirical studies}

\subsection{Setup}

\paragraph{Datasets} The MSCOCO dataset has a standard split: 82,783 images in "train2014" split and 40,504 images in "val2014" split. Each image has 5 human-generated captions in English. We treat the "val2014"  split as our evaluation set and randomly select 1,000 images from ``train2014'' as validation set to be used for early-stopping monitoring. The remaining part in "train2014" which has 408,915 pairs of (image, caption) constructs our training set. For Flickr8K, official split is available leading to 6,000 images for training, 1,000 images for validation and 1,000 images for evaluation. For Flickr30K, we followed the protocol in~\cite{karpathy2014deep} to have 29,000 images for training, 1,014 images for validation and 1,000 images for evaluation.
 
We use Stanford PTBTokenizer~\cite{manning2014stanford} (also used in MSCOCO API), to tokenize the captions in MSCOCO. For Flickr8K and Flickr30K, tokenization are already done by the dataset releaser. Words in the training set are used to construct the vocabulary and those whose frequency less than 20 are discarded. Three special tokens: \#BEGIN\#, \#END\# and \#OOV\# are also taken into consideration, denoting  the starting, the ending of a sentence as well as a universal replacement for out-of-vocabulary words. The final vocabulary sizes are 895, 3,544 and 4,523 for Flickr8K, Flickr30K and MSCOCO respectively.

\paragraph{Objective function}
The objective function of our system is the log likelihood of all the captions given image. $w_{0:t-1}^{(n)}$ represents the previous words before $w_t^{(n)}$. Note that $w_0^{(n)}$ is a special token \#BEGIN\#  inserted before every sentence. $T_n$ is the length of captions $n$.
\vspace{-3mm}
\begin{equation}
\label{equ:condLoss}
L=\frac{1}{N}\sum_n\sum_{t=1}^{T_n} \log p\left(w_t^{(n)}|w_{0:t-1}^{(n)},I^{(n)}\right)
\end{equation}
\vspace{-3mm}

\paragraph{Implementation details} 
We use the ADAM algorithm~\cite{kingma2014adam}, a variant of SGD with adaptive learning rate, to optimize our model. ADAM is advantageous as the  effective step size is invariant to the scale of gradients. This invariance is especially importance to our model as our scene-factorized LSTMs have multiplicative parameters (i.e, $\mA$, $\mB$ and $\mF$) to be optimized jointly.  We didn't use drop-out as we use early stopping as a form of regularization by monitoring the BLEU-1 score on the validation dataset. Ensemble technique is not used in our model.

We observed that a larger minibatch gives a better performance than single sample. Sizes of minibatch in our experiments are 64 for all the three datasets. A typical training session takes about 5 days on a NVidia Titan Z GPU card.

\subsection{Results of image caption on Flickr8K and Flickr30K}

We also validate our system on Flickr8K and Flickr30K by attaining the state-of-the-art results (see Table~\ref{tab:flickrcaption}). As some metrics have not been reported in previous work, we leave them in the table for future comparison.

\begin{table}[t]
\vskip -1em
  \centering
  \small
  \caption{Evaluation of various systems on the task of image captioning}
    \begin{tabular}{|c|ccccccc|}
    \hline
    & BLEU-1 & BLEU-2 & BLEU-3 & BLEU-4 & METEOR & ROUGH-L & CIDEr-D  \\
    \hline
    \multicolumn{8}{|c|}{Flickr8K} \\
    \hline
    DeepVS \cite{karpathy2014deep} & 51 & 31 & 12 & -- & -- & -- & -- \\
    mRNN \cite{mao2014explain} & 58 & 28 & 23 & -- & -- & -- & -- \\
    Google NIC \cite{vinyals2014show} & 63 & 41 & 27 & -- & -- & -- & -- \\
    \textsc{our-(sf+ra)-beam} & \textbf{66.5} & \textbf{47.8} & \textbf{33.2} & \textbf{22.4} & \textbf{20.8} & \textbf{48.6} & \textbf{56.5} \\
    \hline
    \multicolumn{8}{|c|}{Flickr30K} \\
    \hline
    DeepVS \cite{karpathy2014deep} & 50 & 30 & 15 & -- & -- & -- & -- \\
    LRCN \cite{donahue2014long} & 59 & 39 & 25 & 16 & -- & -- & -- \\
    mRNN \cite{mao2014explain} & 60 & 41 & 28 & 19 & -- & -- & -- \\
    Google NIC \cite{vinyals2014show} & \textbf{67} & 45 & 30 & -- & -- & -- & -- \\
    \textsc{our-(sf+ra)-beam} & \textbf{67.0} & \textbf{47.5} & \textbf{33.0} & \textbf{24.3} & \textbf{19.4} & \textbf{47.0} & \textbf{53.1} \\
    \hline
    \end{tabular}
    \label{tab:flickrcaption}

  \end{table}

\subsection{Image and caption retrieval tasks} Retrieving corresponding images (or captions) from given captions (or images) have also been used to evaluate image captioning systems~\cite{karpathy2014deep,kiros2014unifying,mao2014explain} as they are indirectly correlated to the quality of the generated captions.

In particular, the probability of selecting an image given a sentence $P(\mI|\mS_n)$ is proportional to $P(\mS_n|\mI)$ (assuming that a uniform prior of $P(\mI)$, namely, all images occur more or less at the same probability), which is being modeled by image caption systems. Thus, for a training pair $(\mI_n, \mS_n)$, a high-quality image caption system will likely put  $\mI_n$  in the set of the top-ranked retrieved images.  On the other end, for the task of retrieving and ranking captions given images, since there are captions (e.g, those sentences with shorter lengths) having higher probability, using $P(\mS|\mI_n)$ to rank all sentences, given a particular image $\mI_n$ is less likely to be an effective procedure to determine the quality of the captioning system.

\begin{table}[t]
\vskip -1em
  \centering
  \small
   \caption{Evaluation with the tasks of image and captions retrieval on the MSCOCO dataset}
   \label{tab:retrieval}
    \begin{tabular}{|c|cccc|cccc|}
    \hline
    & \multicolumn{4}{c|}{Caption $->$ Image} & \multicolumn{4}{c|}{Image $->$ Caption} \\
    & \textbf{R@1} & \textbf{R@5} & \textbf{R@10} & \textbf{Med} $r$ & \textbf{R@1} & \textbf{R@5} & \textbf{R@10} & \textbf{Med} $r$ \\
    \hline
    \hline
    DeepVS & 20.9 & 52.8 & 69.2 & 4.0 & 29.4 & 62.0 & 75.9 & 2.5 \\
    mRNN & 29.0 & 42.2 & 77.0 & 3.0 & \textbf{41.0} & \textbf{73.0} & \textbf{83.5} & \textbf{2.0} \\
    \textsc{our-ra+sf-beam} & \textbf{29.3} & \textbf{62.8} & \textbf{77.2} & \textbf{2.0} & 36.9 & 67.0 & 78.6 & \textbf{2.0} \\
    \hline
    \end{tabular}
 
\end{table}

Table \ref{tab:retrieval} contrasts our method to several competing ones. Since the retrieval experiment needs to traverse every possible combination of captions and images, only a smaller subset is typically used. Both our system and the compared methods are based on a subset of 1,000 images. The evaluation metrics are the recall rates when returning the top $1$, $5$ or $10$ images (or sentences), as well as the median ranking of the targeted images (or sentences). For the recall rates, the higher the better. For the ranking, the lower the better. On the more relevant task of retrieving images for a given caption, our approach performs the best and improves the state-of-the-art by a large margin. On the less relevant task of retrieving captions, however, our approach is close to or on par with the best approach.

%
\section{Captions generated by our system}
\begin{figure}[t]
\centering
   \includegraphics[width=0.9\columnwidth]{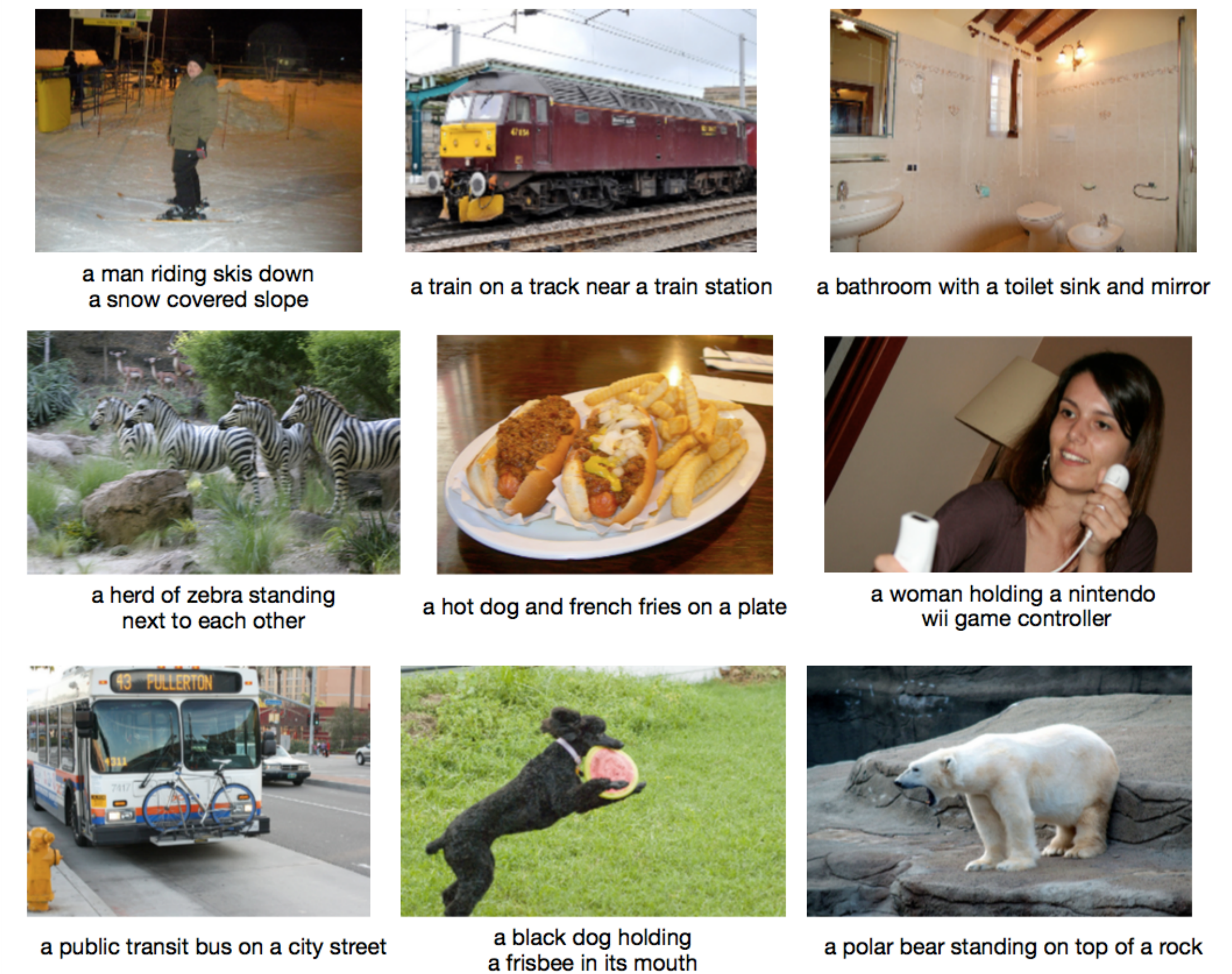}
   \caption{\small Some example captions generated by our system}
\label{fig:example1}
\end{figure}

A group of captions combined with their related images are shown in Fig~\ref{fig:example1}. Our model is able to not only recognize the objects in image but also describe the interaction between objects and scene.

\section{Patch-to-Word Matching}
At any time $t$, our system is presented with the image which is already analyzed and represented with $\cR$ localized patches at multiple scales. We denote the collection of the feature vectors computed from those regions as $\mR = \{\vr_1, \vr_2, \ldots, \vr_{\cR}\}$.

At time $t$, we predict which  visual element is being focused and obtain  the right feature vector as visual context. We use an one-hidden-layer neural network with $\cR$ softmax output variables:
\begin{equation}
p_{it} \propto \exp\,\{ f_v(\vr_i, \mP_{\vw}\vw_{t-1}, \vh_{t-1}, \vv_{t-1})\}, \forall\ i = 1, 2, \cdots, \cR
\end{equation}
where $p_{it}$ denotes the probability of focusing on $i$-th region at time $t$. $f_v(\cdot)$ parameterizes the neural network mapping function (before the softmax). Note that the inputs to the neural network include the extracted features from the $i$-th visual element, and the histories at the previous time step, including the generated word, the abstract meaning  and the visual  context $\vv_{t-1}$. 

To illustrate the transition of the attentions, we compute the weighted sum of pixel values in the image. The weights $p_{it}$ are determined by the amount of attentions predicted for the regions where the pixels belong to. We note that first, the weighted sum of pixel values clearly shows the distinction of ``foreground'' (where the attention would be focusing) and the ``background'' where the weighted sum is smallest (ie, darkest). We also observe that the more focused regions (bordered by red rectangles) correspond well to the concepts in the caption.

The attention weights vary with the word sequence. Fig~\ref{fig:attention} shows the attentions' transition when a sentence goes on. A patch should be allocated high weight when a related word occurs. To further explore the intrinsic semantics in our model, we do a patch-word matching experiment.

\begin{figure}[t]
\begin{center}
\includegraphics[width=0.9\linewidth]{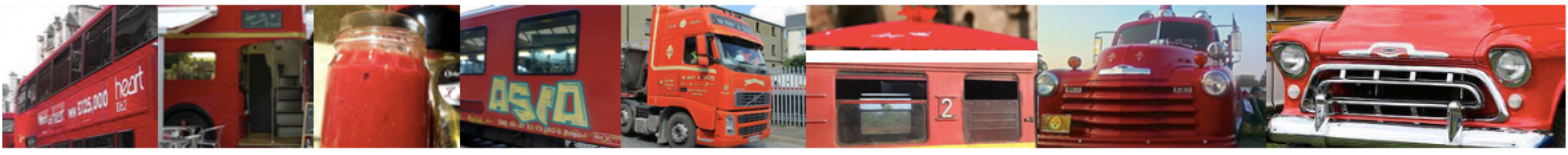}\\
\vskip -0.3em
red\\
\vskip +0.5em
   \includegraphics[width=0.9\linewidth]{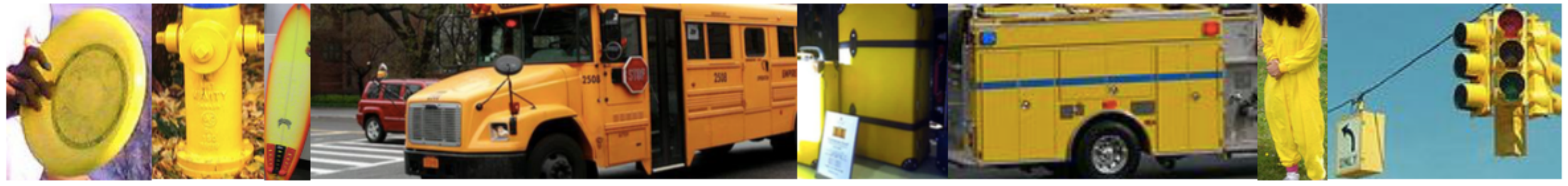}\\
\vskip -0.3em
yellow\\
\vskip +0.5em
\includegraphics[width=0.9\linewidth]{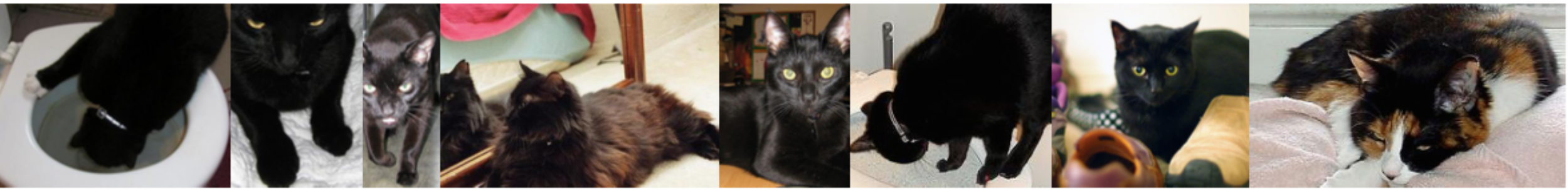}\\
\vskip -0.3em
black cat\\
\vskip +0.5em
   \includegraphics[width=0.9\linewidth]{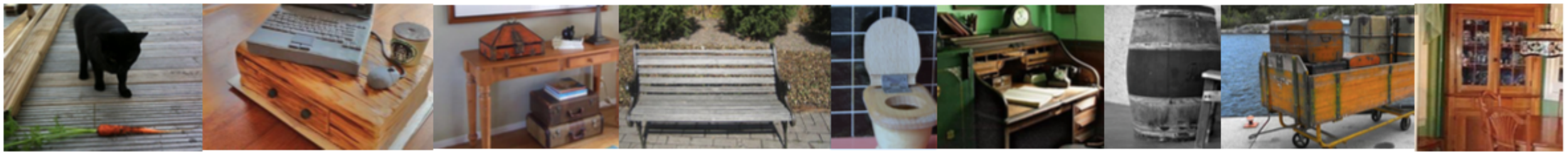}\\
   \vskip -0.3em
wooden\\
\vskip +0.5em
   \includegraphics[width=0.9\linewidth]{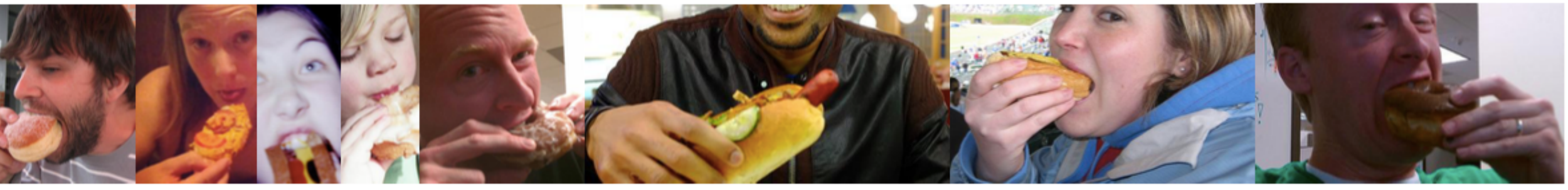}\\
   \vskip -0.3em
bite\\
\vskip +0.5em
\includegraphics[width=0.9\linewidth]{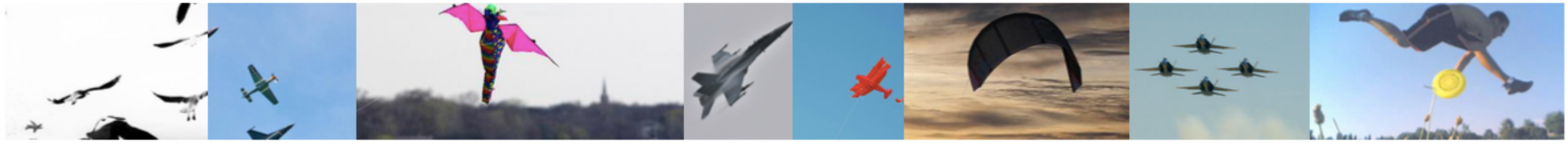}\\
\vskip -0.3em
flying\\
\vskip +0.5em
\includegraphics[width=0.9\linewidth]{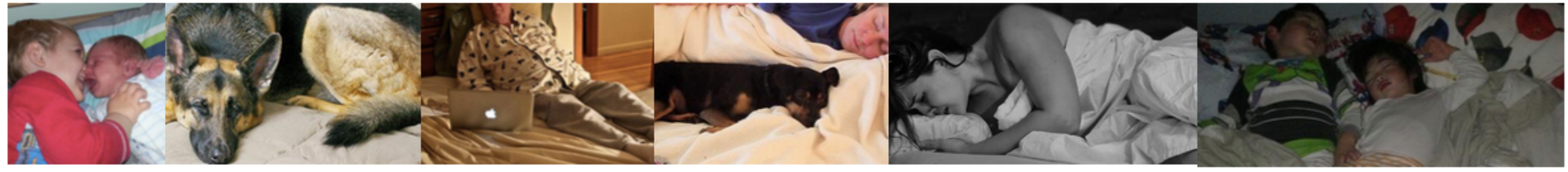}\\
\vskip -0.3em
lying\\
\vskip +0.5em
   \includegraphics[width=0.9\linewidth]{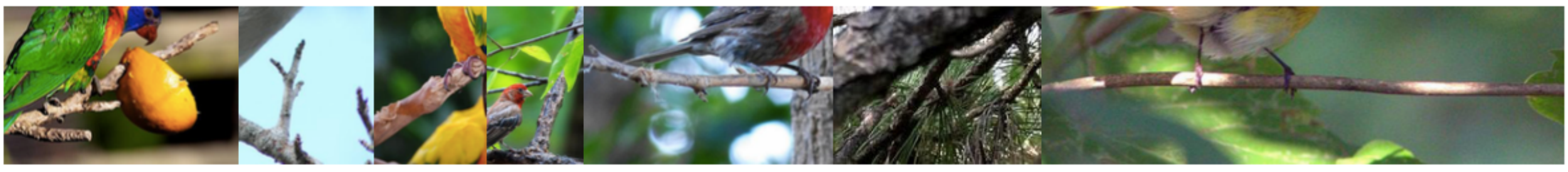}\\
   \vskip -0.3em
branch
\vskip +0.5em
\includegraphics[width=0.9\linewidth]{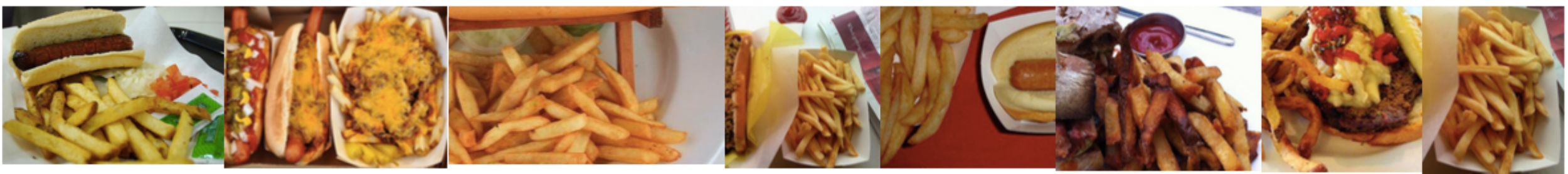}\\
\vskip -0.3em
fries\\
\vskip +0.5em
\includegraphics[width=0.9\linewidth]{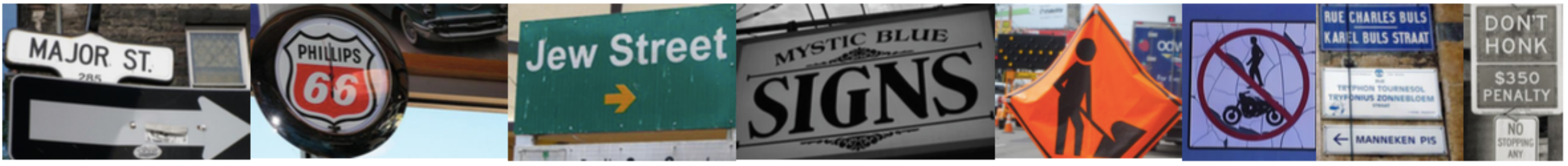}\\
\vskip -0.3em
sign\\
\vskip +0.5em
\includegraphics[width=0.9\linewidth]{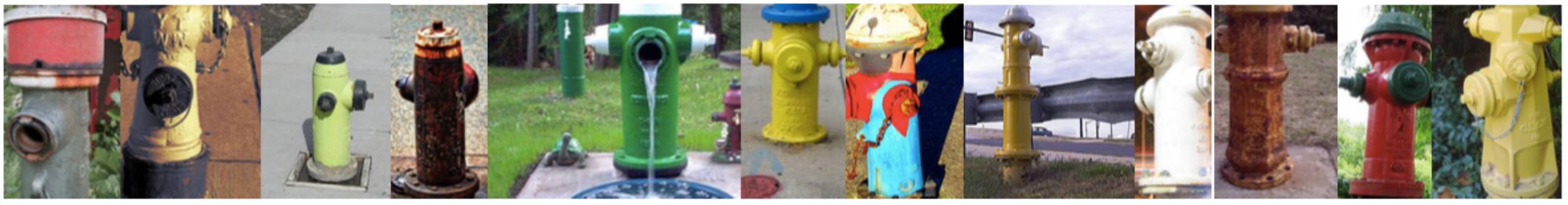}\\
\vskip -0.3em
fire hydrant\\
\vskip +0.5em
\end{center}
   \caption{Matching words to visual regions  for nouns.}
\label{fig:imageWord}
\end{figure}

\begin{figure}[t]
\begin{center}
\includegraphics[width=0.9\linewidth]{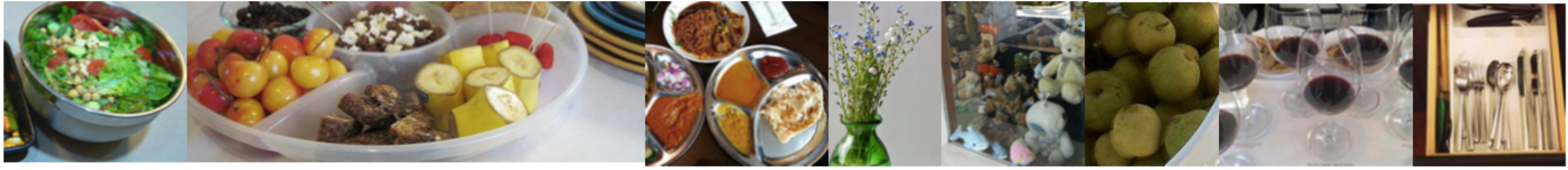}\\
\vskip -0.3em
filled with\\
\vskip +0.5em
\includegraphics[width=0.9\linewidth]{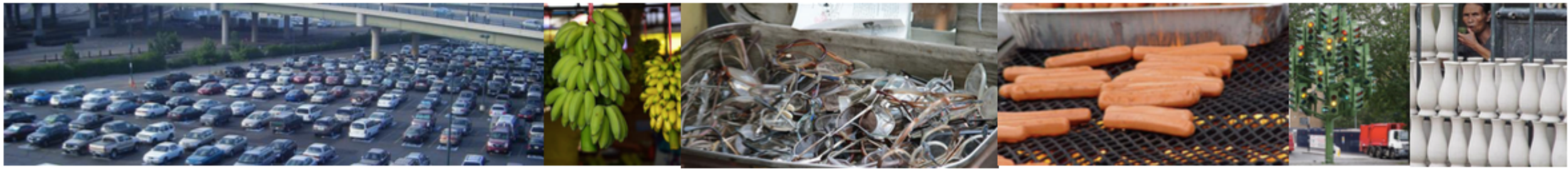}\\
\vskip -0.3em
a bunch of\\
\vskip +0.5em
\includegraphics[width=0.9\linewidth]{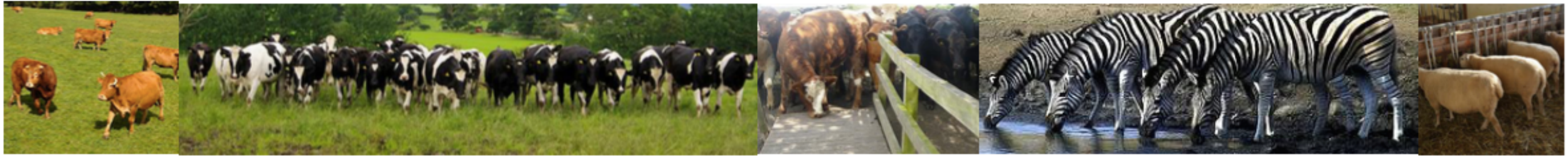}\\
\vskip -0.3em
a herd of\\
\vskip +0.5em
\end{center}
   \caption{Matching words to visual regions  for non-nouns.}
\label{fig:imageWord1}
\end{figure}

Given a word $w$ and a sentence $S$ containing $w$ in time step $t$, we go through the model until time slot $t$ and select the patch with maximum weight at $t$ as the matched patch of word $w$. Under this rule a bunch of patches can be matched to $w$. Fig~\ref{fig:imageWord} and Fig~\ref{fig:imageWord1} shows several words and their matched patches which are randomly chosen from the whole matched patches. Not only the nouns are well learnt to match the objects, but also the verbs and adjectives are learnt to match abstract meaning inside patches. A mapping of semantics in image and text are captured by our model, which is of crucial significance in multi-modal learning.

\begin{figure}[t]
\begin{center}
   \includegraphics[width=0.85\columnwidth]{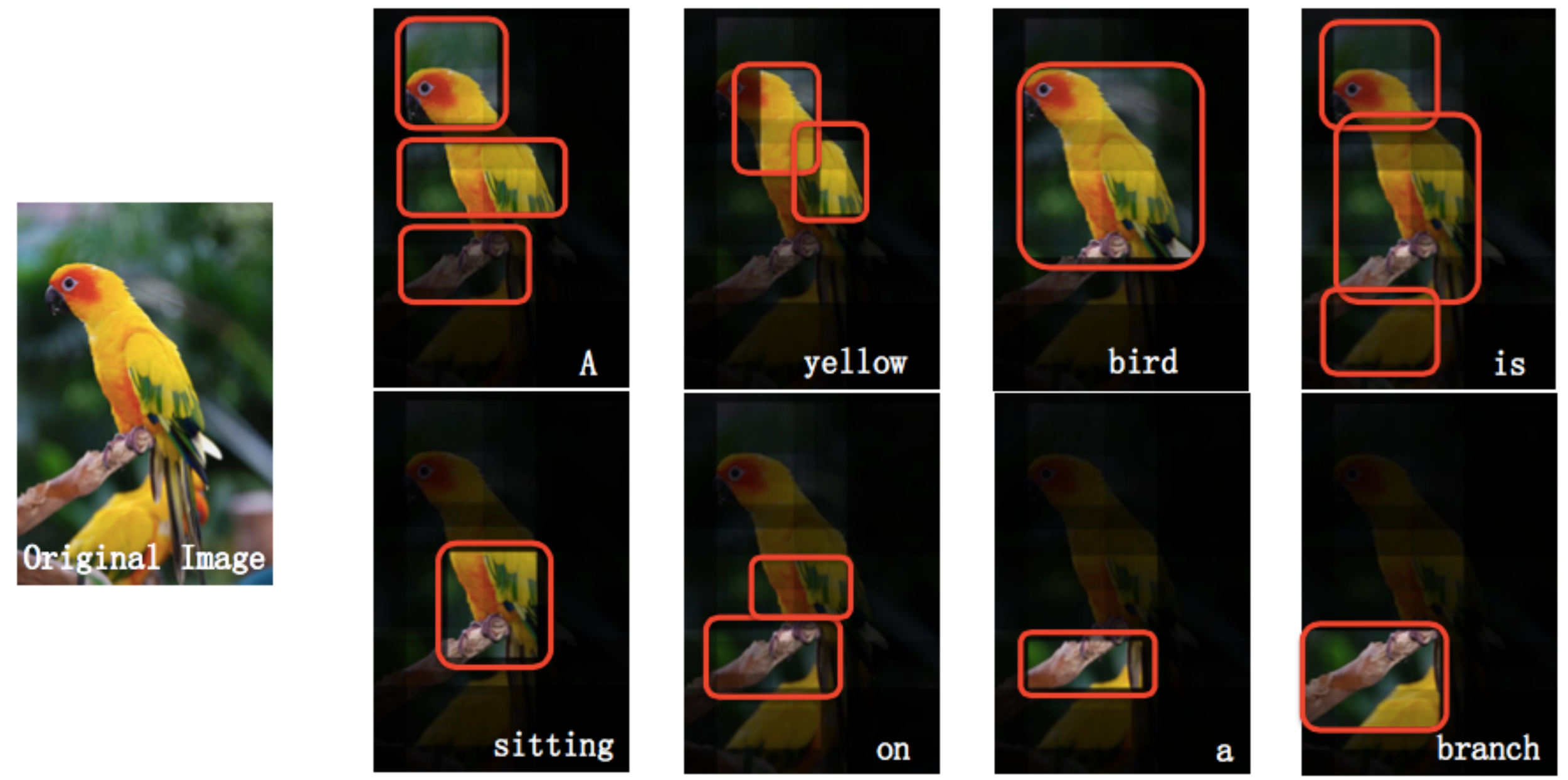}\\
   \includegraphics[width=0.85\columnwidth]{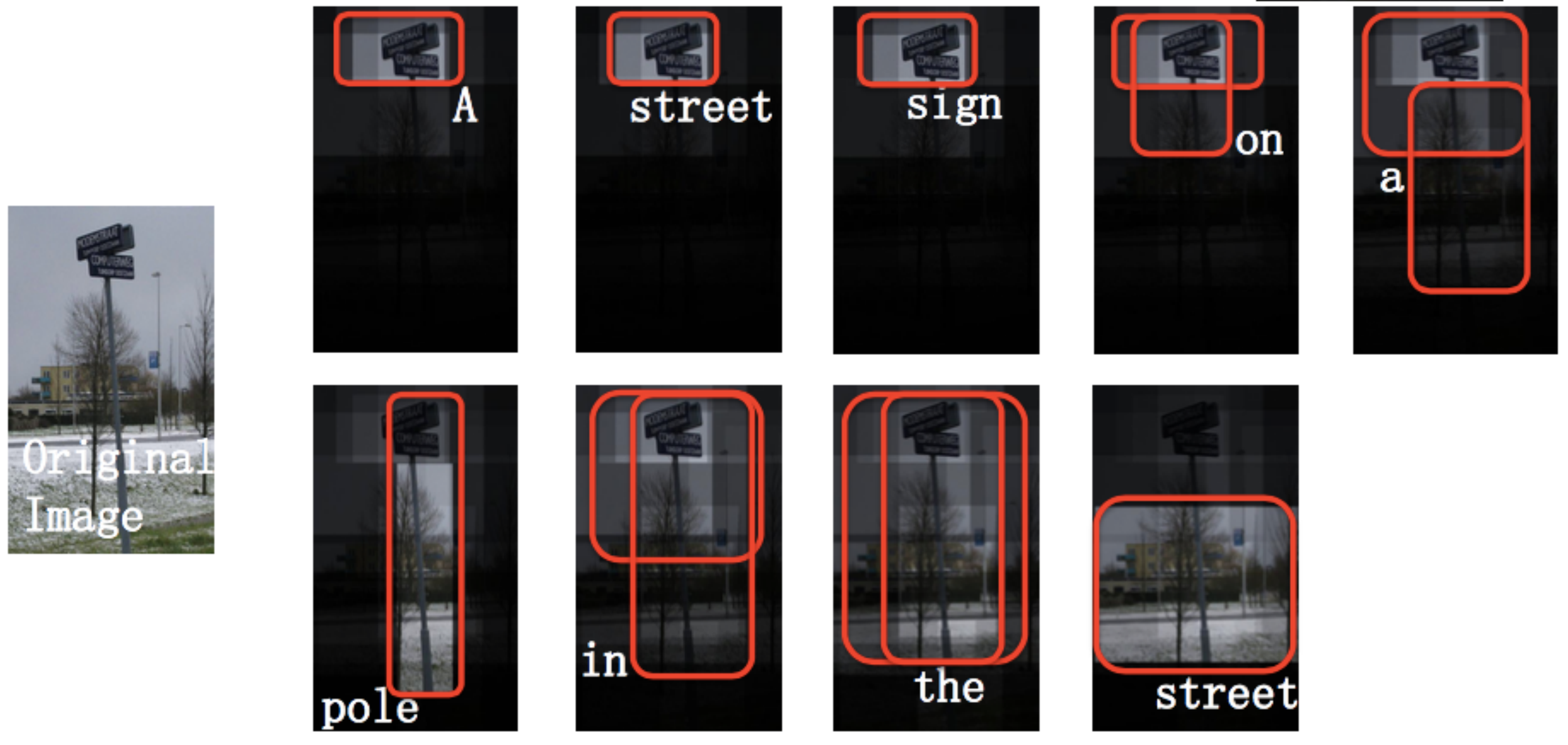}\\
   \includegraphics[width=0.85\columnwidth]{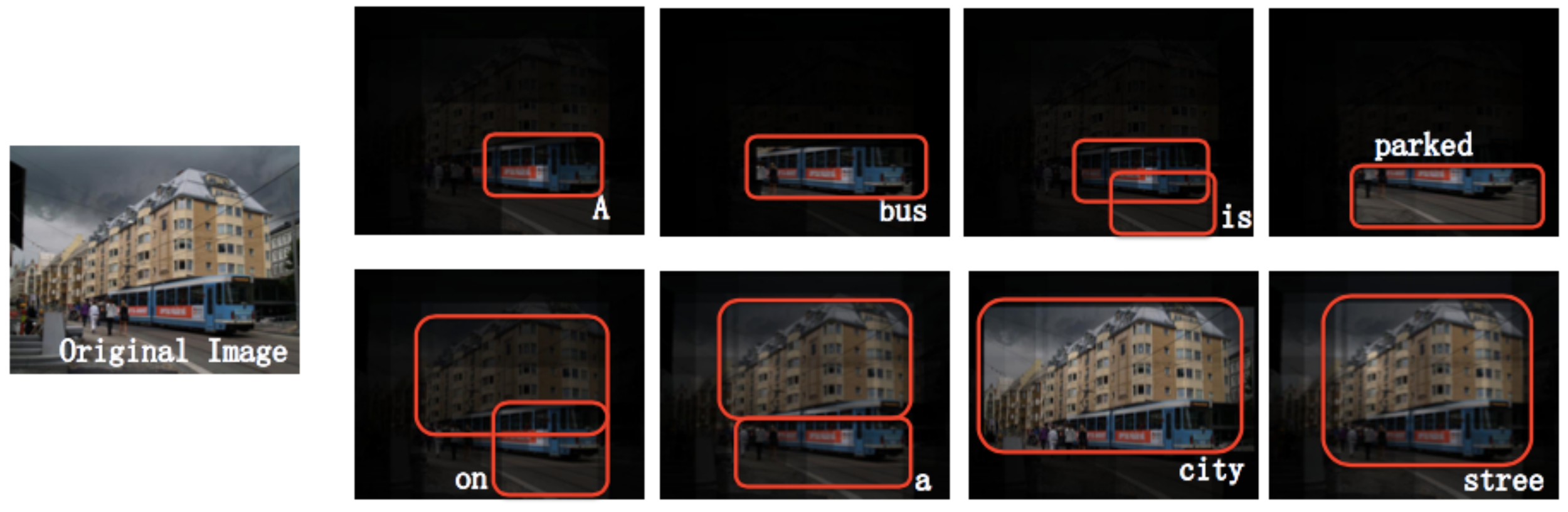}\\
   \includegraphics[width=0.85\columnwidth]{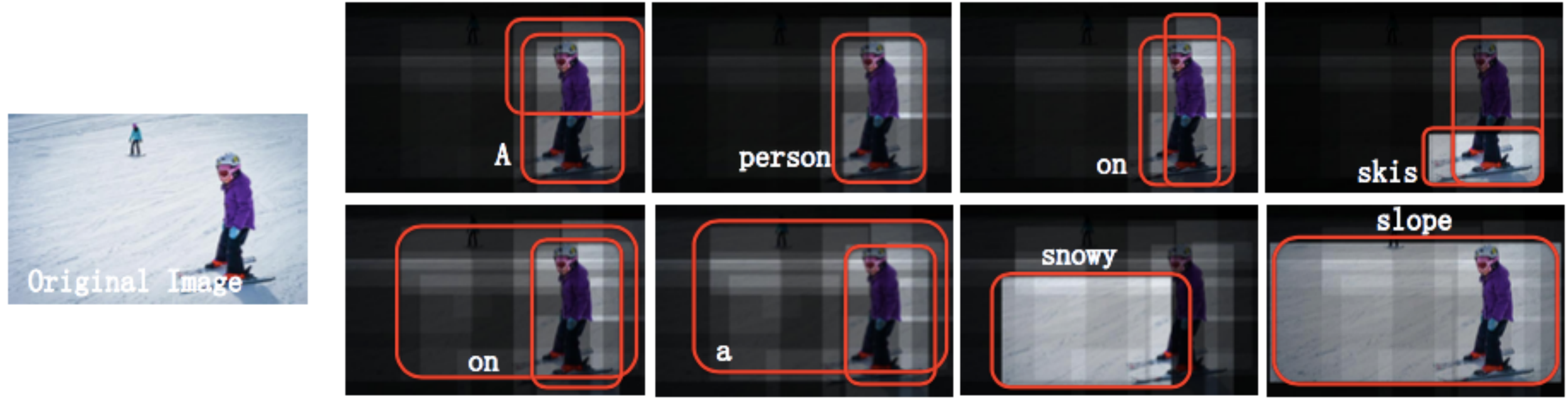}
\end{center}
   \caption{Examples of transitions among visual regions, following the linear orders of the sentences.}
\label{fig:attention}
\end{figure}

\clearpage

\end{document}